\documentclass[10pt,twocolumn,letterpaper]{article}

\usepackage{cvpr}
\usepackage{times}
\usepackage{epsfig}
\usepackage{graphicx}
\usepackage{amsmath}
\usepackage{amssymb}

\usepackage{url}
\usepackage{amsthm}
\usepackage{algorithm}               %format of the algorithm
\usepackage{algorithmic}             %format of the algorithm
\usepackage{multirow}                %multirow for format of table
\usepackage{bm}
\usepackage{footmisc}

\def\1{\mathbf{1}}
\def\0{\mathbf{0}}

\def\A{\bm{A}}
\def\B{\bm{B}}
\def\C{\bm{C}}
\def\D{\bm{D}}
\def\E{\bm{E}}
\def\G{\bm{G}}
\def\N{\bm{N}}

\def\R{\bm{R}}
\def\X{\bm{X}}
\def\Y{\bm{Y}}
\def\Z{\bm{Z}}

\usepackage[pagebackref=true,breaklinks=true,letterpaper=true,colorlinks,bookmarks=false]{hyperref}
 \cvprfinalcopy % *** Uncomment this line for the final submission

\def\cvprPaperID{60} % *** Enter the CVPR Paper ID here

\ifcvprfinal\fi
\begin{document}

%%%%%%%%% TITLE
\title{Multispectral and Hyperspectral Image Fusion by MS/HS Fusion Net}

\author{Qi Xie$^{1}$,  Minghao Zhou$^{1}$, Qian Zhao$^{1}$, Deyu Meng$^{1,}$\footnotemark[1],  ~Wangmeng Zuo$^{2}$, Zongben Xu$^{1}$\\
	$^{1}$Xi'an Jiaotong University; $^{2}$Harbin Institute of Technology\\
	{\tt\small xq.liwu@stu.xjtu.edu.cn~~woshizhouminghao@stu.xjtu.edu.cn~~timmy.zhaoqian@gmail.com}\\  \vspace{-1mm}
	{\tt\small dymeng@mail.xjtu.edu.cn~~wmzuo@hit.edu.cn~~zbxu@mail.xjtu.edu.cn}\\	
	%$^\ast${\small Corresponding author}
}
%\renewcommand{\thefootnote}{\fnsymbol{footnote}}
%\footnotetext[1]{Corresponding author.}
%\renewcommand{\thefootnote}{\arabic{footnote}}
%\thispagestyle{empty}
\maketitle

\renewcommand{\thefootnote}{\fnsymbol{footnote}}
\footnotetext[1]{Corresponding author.}
\renewcommand{\thefootnote}{\arabic{footnote}}
\thispagestyle{empty}

%\thispagestyle{empty}

%%%%%%%%% ABSTRACT
\begin{abstract}\vspace{-3mm}
Hyperspectral imaging can help better understand the characteristics of different materials, compared with traditional image systems. However, only high-resolution multispectral (HrMS)  and low-resolution hyperspectral (LrHS) images can generally be captured at video rate in practice. In this paper, we propose a model-based deep learning approach for merging an HrMS and LrHS images to generate a high-resolution hyperspectral (HrHS) image. In specific, we construct a novel MS/HS fusion model which takes the observation models of low-resolution images and the low-rankness knowledge along the spectral mode of HrHS image into consideration. Then we design an iterative algorithm to solve the model by exploiting the proximal gradient method. And then, by unfolding the designed algorithm, we construct a deep network, called MS/HS Fusion Net, with learning the proximal operators and model parameters by convolutional neural networks.  Experimental results on simulated and real data substantiate the superiority of our method both visually and quantitatively as compared with state-of-the-art methods along this line of research.
\end{abstract}

%%%%%%%%% BODY TEXT
\vspace{-4mm}
\section{Introduction}\vspace{-1mm}

A hyperspectral (HS) image consists of various bands of images of a real scene captured by sensors under different spectrums, which can facilitate a fine delivery of more faithful knowledge under real scenes, as compared to traditional images with only one or a few bands.
The rich spectra of HS images tend to significantly benefit the characterization of the imaged scene and greatly enhance performance in different computer vision tasks, including object recognition, classification, tracking and segmentation \cite{fauvel2013advances,van2010tracking,tarabalka2010segmentation,uzair2013hyperspectral}.%zhang2012comparative

In real  cases, however, due to the limited amount of incident energy, there are critical tradeoffs between spatial and spectral resolution. Specifically, an optical system usually can only provide data with either high spatial resolution but a small number of spectral bands (e.g., the standard RGB image) or with a large number of spectral bands but reduced spatial resolution \cite{michel2011hypxim}.
Therefore, the research issue on merging a high-resolution multispectral (HrMS) image and a low-resolution hyperspectral (LrHS) image to generate a high-resolution hyperspectral (HrHS) image, known as MS/HS fusion, has attracted great attention \cite{yokoya2017hyperspectral}.

  \begin{figure}[t]\vspace{-4mm}
  \begin{center}
     \includegraphics[width=1\linewidth]{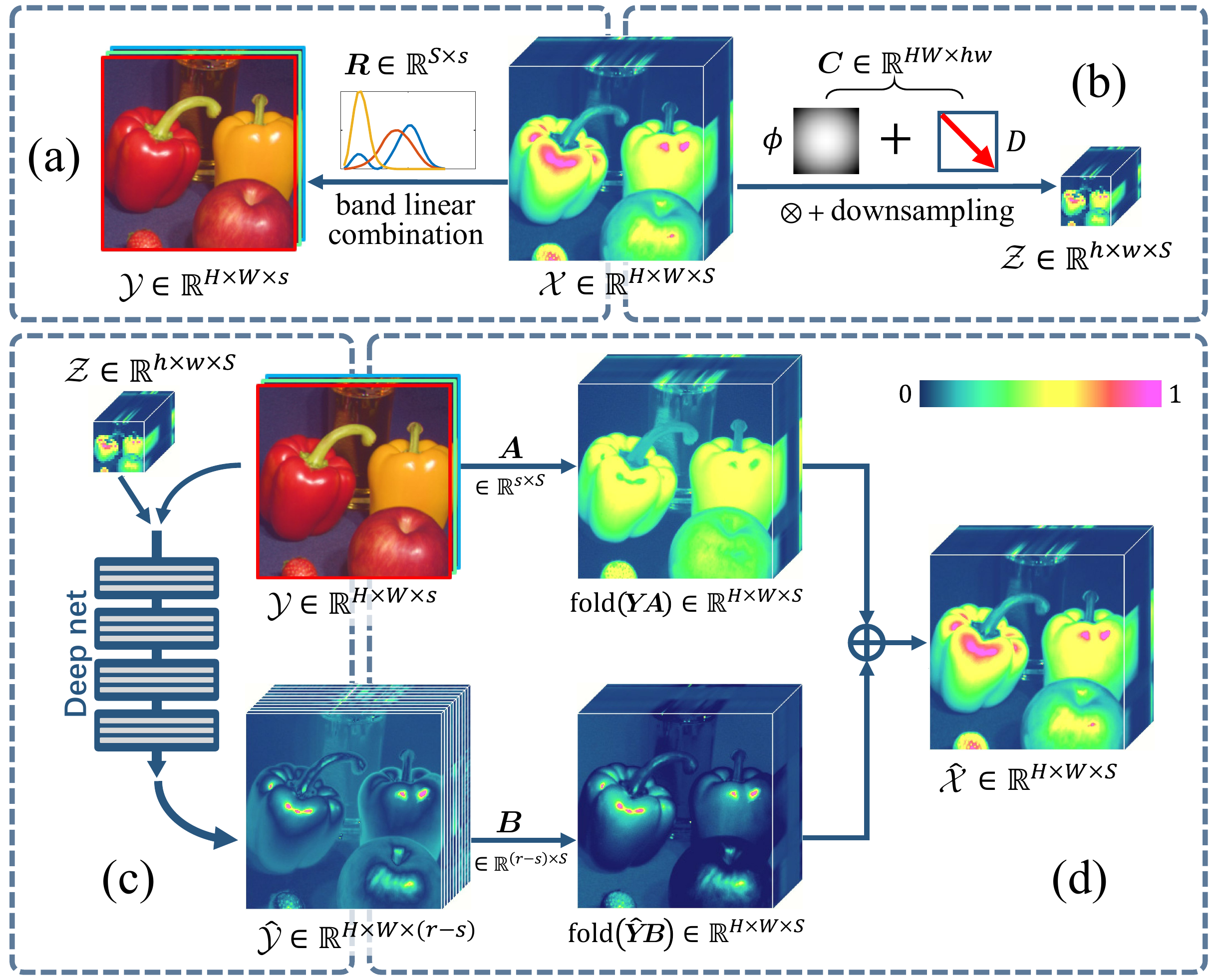}
  \end{center}
  \vspace{-3mm}
     \caption{(a)(b) The observation models for HrMS and LrHS images, respectively. (c) Learning bases $\hat{\Y}$ by deep network, with HrMS $\Y$ and LrHS $\Z$ as the input of the network. (d) The HrHSI $\X$ can be linearly represented by $\Y$  and to-be-estimated $\hat{\Y}$, in a formulation of $\X \approx \Y\A+\hat{\Y}\B$, where  the rank of $\X$ is $r$.   }
  \label{fig:Flowchart}
    \vspace{-4mm}
  \end{figure}

The observation models for the  HrMS and LrHS images are often written as follows \cite{hardie2004map, molina1999bayesian,molina2008variational}:\vspace{-2mm}
\begin{equation}\label{modelY}\vspace{-3mm}
  \Y = \X\R+\N_y,
\end{equation}\vspace{-4mm}
\begin{equation}\label{modelZ}
  \Z = \C\X+\N_z,\vspace{-1mm}
\end{equation}
where $\X\in\mathbb{R}^{HW\times S}$ is the target HrHS image\footnote{The target HS image can also be written as tensor $\mathcal{X}\in \mathbb{\R}^{H\times W\times S}$. We also denote the folding operator for matrix to tensor as: $\mbox{fold}(\X)=\mathcal{X}$.}
with $H$, $W$ and $S$ as its height, width and band number, respectively, $\Y\in\mathbb{R}^{HW\times s}$ is the HrMS image with $s$ as its band number ($s< S$), $\Z\in\mathbb{R}^{hw\times S}$ is the LrHS image with $h$, $w$ and $S$ as its height, width and band number ($h< H$, $w< W$), $\R\in\mathbb{R}^{S\times s}$ is the spectral response of the multispectral sensor as shown in Fig. \ref{fig:Flowchart} (a), $\C\in\mathbb{R}^{hw\times HW}$ is a linear operator which is often assumed to be composed of a cyclic convolution operator $\bm{\phi}$ and a down-sampling matrix $\D$ as shown in Fig. \ref{fig:Flowchart} (b), $\N_y$ and $\N_z$ are the noises contained in HrMS and LrHS images, respectively.
Many methods have been designed based on (\ref{modelY}) and (\ref{modelZ}), and achieved good performance \cite{wei2016blind,huang2014spatial,molina1999bayesian,molina2008variational}.

Since directly recovering the HrHS image $\X$ is an ill-posed inverse problem, many techniques have been exploited to recover $\X$ by assuming certain priors on it. For example, \cite{zhao2011hyperspectral, akhtar2014sparse, grohnfeldt2013jointly} utilize the prior knowledge of HrHS that its spatial information could be sparsely represented under a dictionary trained from HrMS.
Besides, \cite{palsson2014new} assumes the local spatial smoothness prior on the HrHS image and uses total variation regularization to encode it in their optimization model.
Instead of exploring spatial prior knowledge from HrHS, \cite{zhang2015hyperspectral} and \cite{nezhad2016fusion} assume more intrinsic spectral correlation prior on HrHS, and use low-rank techniques to encode such prior along the spectrum to reduce spectral distortions.
Albeit effective for some applications, the rationality of these techniques relies on the subjective prior assumptions imposed on the unknown HrHS to be recovered. An HrHS image collected from real scenes, however, could possess highly diverse configurations both along space and across spectrum. Such conventional learning regimes thus could not always flexibly adapt different HS image structures and still have room for performance improvement.

Methods based on Deep Learning (DL) have outperformed traditional approaches in many computer vision tasks \cite{szegedy2015going} in the past decade, and have been introduced to HS/MS fusion problem very recently~\cite{palsson2017multispectral,scarpa2018target}. As compared with conventional methods, these DL based ones are superior in that they need fewer assumptions on the prior knowledge of the to-be-recovered HrHS, while can be directly trained on a set of paired training data simulating the network inputs (LrHS\&HrMS images) and outputs (HrHS images). The most commonly employed network structures include CNN \cite{dong2016image}, 3D CNN \cite{palsson2017multispectral}, and residual net \cite{scarpa2018target}.
%The network input is generally attained by stacking an HrMS image $\Y$ and an LrHS image $\Z$ together ($\Z$ is usually  interpolated to the same spatial size as $\Y$ in advance), and all network parameters are updated directly using Stochastic gradient descent (SGD) or other training algorithms.
Like other image restoration tasks where DL is successfully applied to, these DL-based methods have also achieved good resolution performance for MS/MS fusion task.

%the main drawback of this technique is the lack of interpretability
However, the current DL-based MS/HS fusion methods still have evident drawbacks. The most critical one is that these methods use general frameworks for other tasks, which are not specifically designed for MS/HS fusion. This makes them lack interpretability specific to the problem. In particular, they totally neglect the observation models (\ref{modelY}) and (\ref{modelZ}) \cite{palsson2017multispectral,scarpa2018target}, especially the operators $\R$ and $\C$, which facilitate an understanding of how LrHS and HrMs are generated from the HrHS. Such understanding, however, should be useful for calculating HrHS images. Besides this generalization issue, current DL methods also neglect the general prior structures of HS images, such as spectral low-rankness. Such priors are intrinsically possessed by all meaningful HS images,
%and the neglect of such priors makes DL-based methods go to another extreme as conventional methods, and the potential capacity thus still has room to be further enhanced.
%and the neglect of such priors makes DL-based methods still has room to be further enhanced.
and the neglect of such priors implies that DL-based methods still have room for further enhancement.

%Such priors are intrinsically possessed by almost all meaningful HS images, which implies that the potential capacity of current DL-based MS/HS fusion methods still has room to be further enhanced.

%Other methods based on component substitution\cite{shettigara1992generalized}, 3D-wavelet transform \cite{zhang2007multi}, total variation \cite{palsson2014new} also are also proposed and achieve good performance.

In this paper, we propose a novel deep learning-based method that integrates the observation models and image prior learning into a single network architecture.
This work mainly contains the following three-fold contributions:

Firstly, we propose a novel MS/HS fusion model, which not only takes the observation models (\ref{modelY}) and (\ref{modelZ}) into consideration but also exploits the approximate low-rankness prior structure along the spectral mode of the HrHS image to reduce spectral distortions \cite{zhang2015hyperspectral,nezhad2016fusion}.
Specifically, we prove that if and only if observation model (\ref{modelY}) can be satisfied,
the matrix of HrHS image $\X$ can be linearly represented by the columns in HrMS matrix $\Y$ and a to-be-estimated matrix $\hat{\Y}$, i.e., $\X = \Y\A+\hat{\Y}\B$ with coefficient matrices $\A$ and $\B$. One can see Fig. \ref{fig:Flowchart} (d) for easy understanding.
We then construct a concise model by combining the observation model (\ref{modelZ}) and the linear representation of $\X$. We also exploit the proximal gradient method \cite{beck2009fast} to design an iterative algorithm to solve the proposed model.

Secondly, we unfold this iterative algorithm into a deep network architecture, called MS/HS Fusion Net or MHF-net, to implicitly learn the to-be-estimated $\hat{\Y}$, as shown in Fig. \ref{fig:Flowchart} (c).  After  obtaining $\hat{\Y}$, we can then easily achieve $\X$ with $\Y$ and $\hat{\Y}$. To the best of our knowledge, this is the first deep-learning-based MS/HS fusion method that fully considers the intrinsic mechanism of the MS/HS fusion problem. Moreover, all the parameters involved in the model can be automatically learned from training data in an end-to-end manner. This  means that the spatial and spectral responses ($\R$ and $\C$) no longer need to be estimated beforehand as most of the traditional non-DL methods did, nor to be fully neglected as current DL methods did.

%Experiments on simulated and real noisy data sets demonstrate the superiority of the proposed MHF-net as compared with the stat-of-the-art methods along this line.

Thirdly, we have collected or realized current state-of-the-art algorithms for the investigated MS/HS fusion task, and compared their performance on a series of synthetic and real problems. The experimental results comprehensively substantiate the superiority of the proposed method, both quantitatively and visually.

%\section{Notations and preliminaries}
In this paper, we denote scalar, vector, matrix and tensor in non-bold case, bold lower case, bold upper case and calligraphic upper case letters, respectively.

\vspace{-2mm}
\section{Related work}
\vspace{-1mm}
%Approaches for MS/HS fusion can be generally grouped into two categories: traditional methods and the learning based methods.
\subsection{Traditional methods}\vspace{-1mm}
The pansharpening technique in remote sensing is closely related to the investigated MS/HS problem. This task aims to obtain a high spatial resolution MS image by the fusion of a MS image and a wide-band panchromatic image. A heuristic approach to perform MS/HS fusion is  to treat it as a number of pansharpening sub-problems, where each band of the HrMS image plays the role of a panchromatic image.
There are mainly two categories of pansharpening methods: component substitution (CS) \cite{chavez1991comparison,laben2000process, aiazzi2007improving}
and multiresolution analysis (MRA) \cite{loncan2015hyperspectral, mallat1989theory,burt1987laplacian,starck2007undecimated,do2005contourlet}.
%The typical CS-based pansharpening methods including principal component analysis \cite{chavez1991comparison} and Gramm-Schmidt orthogonalization \cite{laben2000process,aiazzi2007improving}. The MRA-based approaches inject spatial details of the panchromatic image into the MS image, with exploiting multiscale decomposition methods, such as decimated wavelet transform \cite{mallat1989theory}, Laplacian pyramid \cite{burt1987laplacian}, curvelets \cite{starck2007undecimated,do2005contourlet}, etc.
These methods always suffer from the high spectral distortion, since a single panchromatic image contains little spectral information as compared with the expected HS image.

In the last few years, machine learning based methods have gained much attention on MS/HS fusion problem \cite{zhao2011hyperspectral, akhtar2014sparse, grohnfeldt2013jointly,huang2014spatial,zhang2015hyperspectral,yokoya2011coupled,nezhad2016fusion,wei2016blind}.
Some of these methods used sparse coding technique  to learn a dictionary on the patches across a HrMS image, which delivers spatial knowledge of HrHS to a certain extent, and then learn a coefficient matrix from LrHS to fully represent the HrHS \cite{zhao2011hyperspectral, akhtar2014sparse, grohnfeldt2013jointly,wei2016blind}. Some other methods, such as \cite{huang2014spatial}, use the sparse matrix factorization to learn a spectral dictionary for LrHS images and then construct HrMS images by exploiting both the spectral dictionary and HrMS images. The low-rankness of HS images can also be exploited with non-negative matrix factorization, which helps to reduce spectral distortions and enhances the MS/HS fusion performance \cite{zhang2015hyperspectral,yokoya2011coupled,nezhad2016fusion}. The main drawback of these methods is that they are mainly designed based on human observations and strong prior assumptions, which may not be very accurate and would not always hold for diverse real world images.

%The total variation based method  is also used to encode the local smoothness prior of the HS image in \cite{palsson2014new}. \cite{nezhad2016fusion} uses spectral mixing model to reduce spectral distortions. \cite{zhang2015hyperspectral} propose a method using non-negative matrix factorization to encode the low-rankness of the HS image. However, these techniques are mainly designed according to human observations and would not always hold for diverse real world images and thus can hardly further enhance the recovery performance.

\subsection{Deep learning based methods}\vspace{-1mm}
Recently, a number of DL-based pansharpening methods were proposed by exploiting different network structures \cite{huang2015new,masi2016pansharpening,wei2017deep,wei2017boosting,rao2017residual,scarpa2018target,shao2018remote}. These methods can be easily adapted to MS/HS fusion problem. For example, very recently, \cite{palsson2017multispectral} proposed a 3D-CNN based MS/HS fusion method by using PCA to reduce the computational cost.  This method is usually trained with prepared training data. The network inputs are set as the combination of HrMS/panchromatic images and LrHS/multispectral images (which is usually  interpolated to the same spatial size as HrMS/panchromatic images in advance), and the outputs are the corresponding HrHS images. The current DL-based methods have been verified to be able to attain good performance. They, however, just employ networks assembled with some off-the-shelf components in current deep learning toolkits, which are not specifically designed against the investigated problem. Thus the main drawback of this technique is the lack of interpretability to this particular MS/HS fusion task. In specific, both the intrinsic observation model (\ref{modelY}), (\ref{modelZ}) and the evident prior structures, like the spectral correlation property, possessed by HS images have been neglected by such kinds of ``black-box" deep model.
% which makes it still have a large room for further capability enhancement.

\vspace{-2mm}
\section{MS/HS fusion model}
\vspace{-1mm}
In this section, we demonstrate the proposed MS/HS fusion model in detail.
\subsection{Model formulation}\vspace{-1mm}
We first introduce an equivalent formulation for observation model (\ref{modelY}). Specifically, we have following theorem\footnote{All proofs are presented in supplementary material.}.
\newtheorem{Thm}{Theorem}
\begin{Thm}
For any $\X\in\mathbb{R}^{HW\times S}$ and $\tilde{\Y}\in\mathbb{R}^{HW\times s}$, if $\mbox{\emph{rank}}(\X)=r>s$ and  $\mbox{\emph{rank}}(\tilde{\Y})=s$, then the following two statements are equivalent to each other:\\
(a) There exists an $\R\in \mathbb{R}^{S\times s}$, subject to,
\begin{equation}\label{assumption1}
  \tilde{\Y} = \X\R.
\end{equation}
(b) There exist $\A\in \mathbb{R}^{s\times S}$, $\B\in \mathbb{R}^{(r-s)\times S}$ and $\hat{\Y}\in \mathbb{R}^{HW\times (r-s)}$, subject to,
\begin{equation}\label{assumption2}
  \X = \tilde{\Y}\A + \hat{\Y}\B.
\end{equation}
\end{Thm}

In reality, the band number of an HrMS image is usually not large, which makes it full rank along spectral mode.
For example,  the most commonly used HrMS images, RGB images, contain three bands, and their rank along the spectral mode is usually also three.
Thus, by letting $\tilde{\Y} = \Y-\N_y$ where $\Y$ is the observed HrMS in (\ref{modelY}), it is easy to find that $\tilde{\Y}$ and $\X$ satisfy the conditions in \textbf{Theorem 1}. Then the observation model (\ref{modelY}) is equivalent to
\begin{equation}\label{modelX}
  \X = \Y\A + \hat{\Y}\B+\N_x,
\end{equation}
where $\N_x =-\N_y\A$ is caused by the noise contained in the HrMS image. In (\ref{modelX}), $[{\Y},\hat{\Y}]$ can be viewed as $r$ bases that represent columns in $\X$ with coefficients matrix $\left[\A;\B\right]\in\mathbb{R}^{r\times S}$, where only the $r-s$ bases in $\hat{\Y}$ are unknown. In addition, we can derive the following corollary:
\newtheorem{Coro}{Corollary}
\begin{Coro}
For any  $\tilde{\Y}\in\mathbb{R}^{HW\times s}$,  $\tilde{\Z}\in\mathbb{R}^{hw\times S}$, $\C\in\mathbb\mathbb{R}^{hw\times HW}$, if $\mbox{\emph{rank}}(\tilde{\Y})=s$ and $\mbox{\emph{rank}}(\tilde{\Z})=r>s$, then the following two statements are equivalent to each other:\\
(a) There exist $\X\in\mathbb{R}^{HW\times S}$  and  $\R\in \mathbb{R}^{S\times s}$, subject to,
\begin{equation}\label{assumption3}
  \tilde{\Y} = \X\R, ~~
  \tilde{\Z} = \C\X,~~ \mbox{\emph{rank}}(\X)=r.
\end{equation}
(b) There exist  $\A\in \mathbb{R}^{s\times S}$, $r>s$, $\B\in \mathbb{R}^{(r-s)\times S}$ and $\hat{\Y}\in \mathbb{R}^{HW\times (r-s)}$, subject to,
\begin{equation}\label{assumption4}
  \tilde{\Z} = \C\left(\tilde{\Y}\A + \hat{\Y}\B\right).
\end{equation}
\end{Coro}
By letting $\tilde{\Z}=\Z-N_z$, it is easy to find that, when being viewed as equations of the to-be-estimated $\X$, $\R$ and $\C$, the observation model (\ref{modelY}) and model (\ref{modelZ}) are equivalent to the following equation of $\hat{\Y}$, $\A$, $\B$  and $\C$:
\begin{equation}\label{modelAll}
  \Z = \C\left(\Y\A + \hat{\Y}\B\right) + \N,
\end{equation}
where $\N = \N_z-\C\N_y\A$ denotes the noise contained in HrMS and LrHS image.

By (\ref{modelAll}), we design the following MS/HS fusion model:
\begin{equation}\label{Problem}
  \min_{\hat{\Y}}\left\|\C\left(\Y\A+\hat{\Y}\B\right)-\Z\right\|_F^2+\lambda f\left(\hat{\Y}\right),
\end{equation}
where $\lambda$ is a trade-off parameter, and $f(\cdot)$ is a regularization function. We adopt regularization on the to-be-estimated bases in $\hat{\Y}$, rather than on $\X$ as in traditional methods.
This will help alleviate destruction of the spatial detail information in the known $\Y$\footnote{Many regularization terms, such as total variation norm, will lead to loss of details like the sharp edge, lines and high light point in the image.} when representing $\X$ with it.

It should be noted that for the same data set, the matrices $\A$, $\B$ and $\C$ are fixed. This means that these matrices can be learned from the training data. In the later sections we will show how to learn them with a deep network.

\subsection{Model optimization}\vspace{-1mm}
We now solve (\ref{Problem}) using a proximal gradient algorithm \cite{beck2009fast}, which iteratively updates $\hat{\Y}$ by calculating
\begin{equation}\label{subProblem}
  \hat{\Y}^{(k+1)} = \arg\min_{\hat{\Y}}Q\left(\hat{\Y},\hat{\Y}^{(k)}\right),
\end{equation}
where $\hat{\Y}^{(k)}$  is the updating result after $k-1$ iterations, $k=1,2,\cdots,K$, and $Q(\hat{\Y},\hat{\Y}^{(k)})$ is a quadratic approximation \cite{beck2009fast} defined as:
\begin{equation}\label{Qproblem}
\begin{split}
Q\!\left(\!\hat{\Y},\hat{\Y}^{(k)}\!\right)& \!=\! g\left(\!\hat{\Y}^{(k)}\!\right) \!+\!\left\langle \hat{\Y}-\hat{\Y}^{(k)}, \nabla g\left(\!\hat{\Y}^{(k)}\!\right) \right\rangle \\
&+ \frac{1}{2\eta}\left\| \hat{\Y}-\hat{\Y}^{(k)} \right\|_F^2 + \lambda f\left( \hat{\Y} \right),
\end{split}
\end{equation}
where $g(\hat{\Y}^{(k)}) = \|\C(\Y\A+\hat{\Y}^{(k)}\B)-\Z\|_F^2$ and $\eta$ plays the role of stepsize.

It is easy to prove that the  problem (\ref{subProblem}) is equivalent to:
\begin{equation}\label{subProblem2}
  \min_{\hat{\Y}}\frac{1}{2}\left\| \hat{\Y} \!-\! \left(\! \hat{\Y}^{(k)} \!-\! \eta\nabla g\left(\hat{\Y}^{(k)}\right)  \!\right) \right\|_F^2 \!\!+\! \lambda\eta f\left( \hat{\Y} \right).
\end{equation}
For many kinds of regularization terms, the solution of  Eq. (\ref{subProblem2}) is usually in a closed-form \cite{donoho1995noising}, written as:
\begin{equation}\label{CloseF}
  \hat{\Y}^{(k+1)} = \mbox{prox}_{\lambda\eta}\left(\! \hat{\Y}^{(k)} \!-\! \eta\nabla g\left(\hat{\Y}^{(k)}\right)  \!\right).
\end{equation}
%By plugging\vspace{-1mm}
%\begin{equation}\label{G}
%  \nabla g\left(\!\hat{\Y}^{(k)}\!\right) = \C^T\left(\C\left(\Y\A+\hat{\Y}^{(k)}\B\right)-\Z\right)\B^T,
%\vspace{-1mm}
%\end{equation}
Since  $
  \nabla g\left(\!\hat{\Y}^{(k)}\!\right) = \C^T\!\left(\C\left(\Y\A\!+\!\hat{\Y}^{(k)}\B\right)\!-\!\Z\right)\!\B^T$,
we can obtain the final updating rule for $\hat{\Y}$:
\small
\begin{equation}\label{updateY}
  \hat{\Y}^{(k+1)} \!=\!
  \mbox{prox}_{\lambda\eta}\!\!\left(\! \hat{\Y}^{(k)}
  \!-\! \eta\C^T\!\left(\C\!\left(\Y\!\A+\hat{\Y}^{(k)}\!\B\right)\!-\!\Z\right)\!\B^T  \!\right)\!. \vspace{-0.5mm}
\end{equation}
\normalsize
In the later section, we will unfold this algorithm into a deep network.

\vspace{-2mm}
\section{MS/HS fusion net}\vspace{-1mm}
Based on the above algorithm, we build a deep neural network for MS/HS fusion by unfolding all steps of the algorithm as network layers. This technique has been widely utilized in various computer vision tasks and has been substantiated to be effective in compressed sensing, dehazing, deconvolution, etc. \cite{yang2018proximal, yang2017admm, zhang132017learning}.
The proposed network is a structure of $K$ stages implementing $K$ iterations in the iterative algorithm for solving Eq. (\ref{Problem}), as shown in Fig. \ref{fig:Network} (a) and (b). Each stage takes the HrMS image $\Y$, LrHS image $\Z$, and the output of the previous stage $\hat{\Y}$, as inputs, and outputs an updated $\hat{\Y}$ to be the new input of next layer.

  \begin{figure}[t]
  \begin{center}
  \vspace{-2mm}
     \includegraphics[width=1\linewidth]{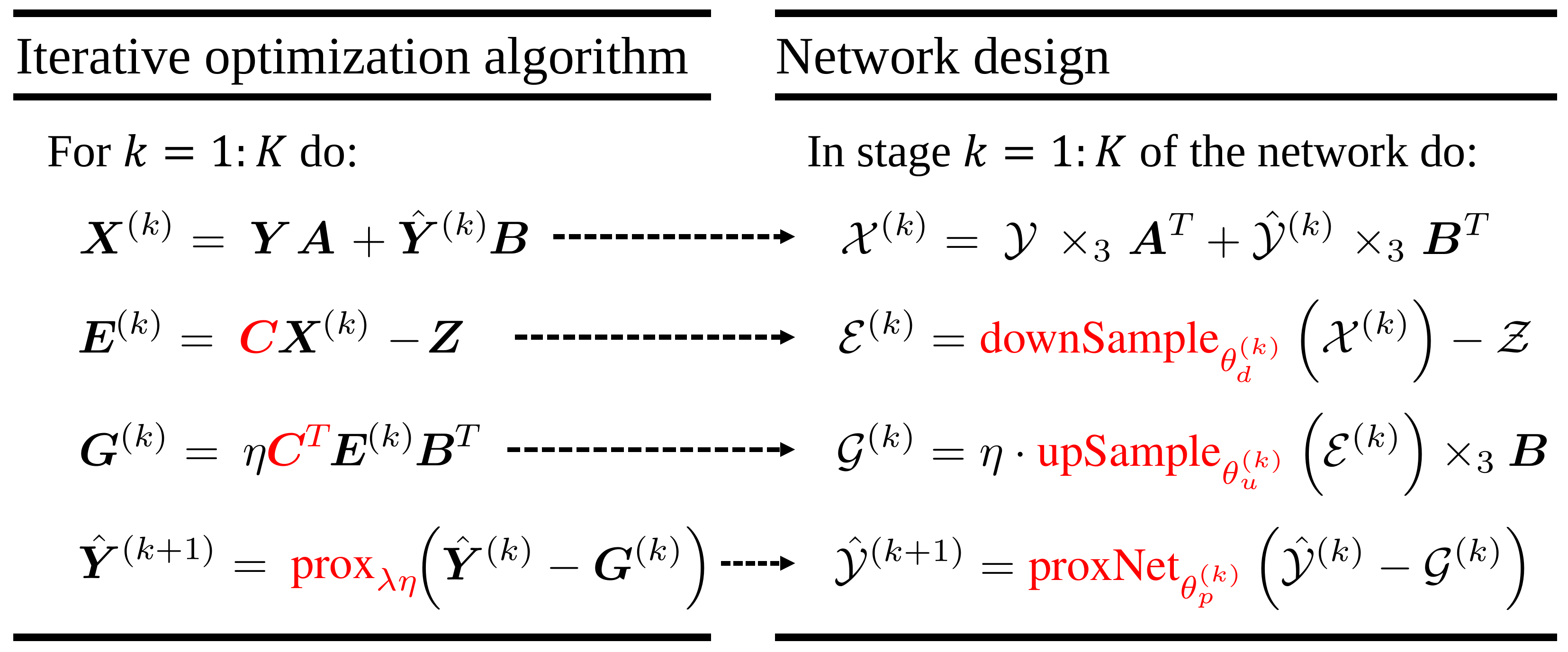}
  \end{center}
  \vspace{-4mm}
     \caption{An illustration of relationship between the algorithm with matrix form and the network structure with tensor form.}
  \label{fig:Algorithm}
    \vspace{-3mm}
  \end{figure}

  \begin{figure*}[t]
  \begin{center}
  \vspace{-4.5mm}
     \includegraphics[width=0.96\linewidth]{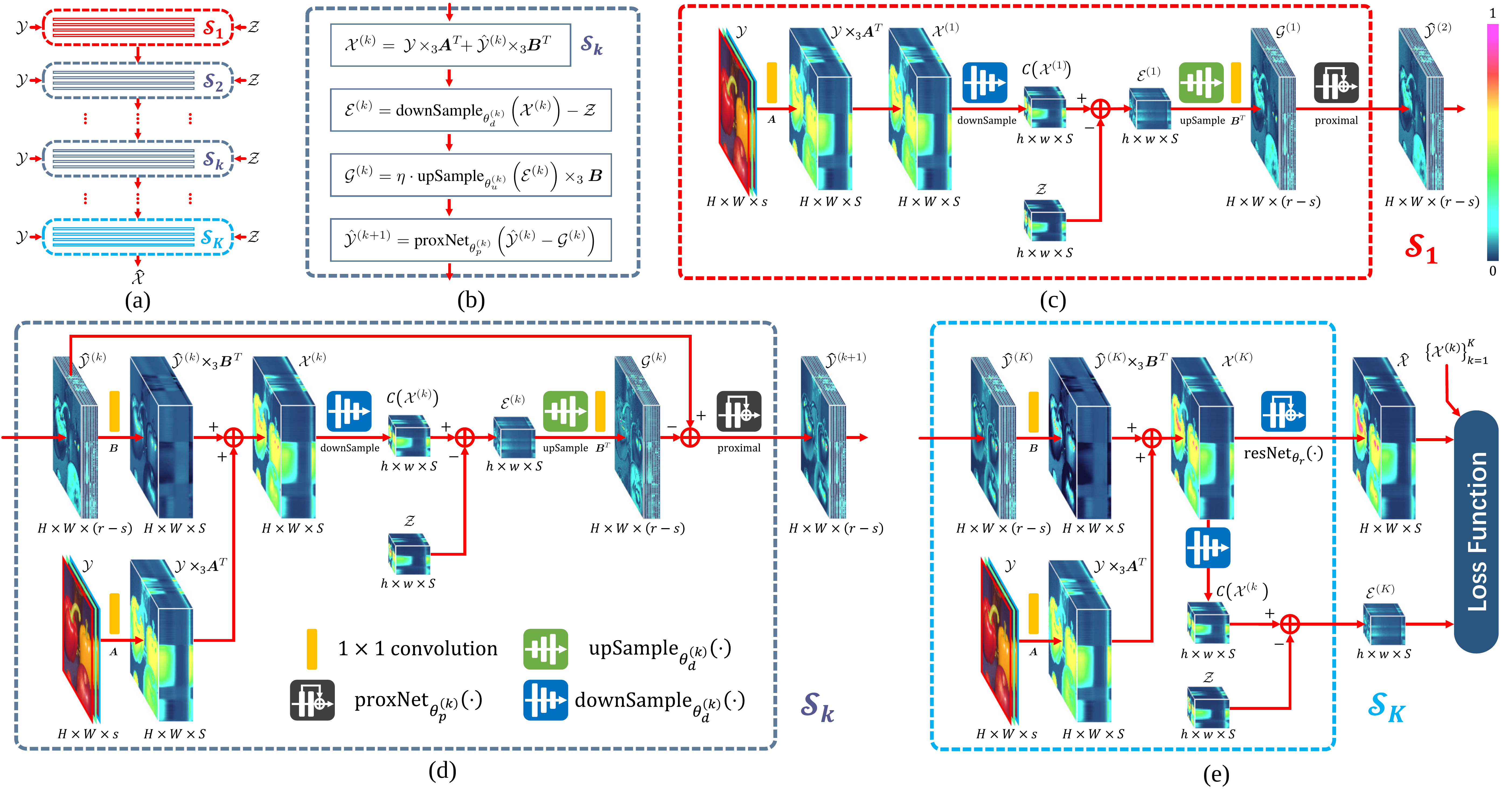}
  \end{center}
  \vspace{-4.7mm}
     \caption{(a) The proposed network with $K$ stages implementing $K$ iterations in the iterative optimization algorithm, where the $k^{th}$ stage is denoted as ${\mathcal{S}_k}, (k = 1,2,\cdots,K)$. (b) The flowchart of $k^{th}$ ($k<K$) stage.  (c)-(e) Illustration of the first, $k^{th}$ ($1<k<K$) and final stage of the proposed network, respectively.  When setting $\hat{\mathcal{Y}}^{(k)}=\bm{0}$, ${\mathcal{S}_k}$ is equivalent to ${\mathcal{S}_1}$.}
  \label{fig:Network}
    \vspace{-3.9mm}
  \end{figure*}

\subsection{Network design}\vspace{-1mm}

\textbf{Algorithm unfolding.} We first decompose the updating rule (\ref{updateY}) into the following four sequential parts:
\begin{equation}\label{theX}
  \X^{(k)} = \Y\A+\hat{\Y}^{(k)}\B,
\end{equation}
\begin{equation}\label{theE}
  \E^{(k)} = \C\X^{(k)}-\Z,
\end{equation}
\begin{equation}\label{theG}
  \G^{(k)} = \eta\C^T\E^{(k)}\B^T,
\end{equation}
\begin{equation}\label{theY}
  \hat{\Y}^{(k+1)} = \mbox{prox}_{\lambda\eta}\left(\hat{\Y}^{(k)} - \G^{(k)}\right).
\end{equation}
In the network framework, we use the images with their tensor formulations ($\mathcal{X}\in\mathbb{R}^{H\times W \times S}$, $\mathcal{Y}\in\mathbb{R}^{H\times W \times s}$ and $\mathcal{Z}\in\mathbb{R}^{h\times w \times S}$) instead of their matrix forms to protect their original structure knowledge and make the network structure (in tensor form) easily designed. We then design a network to approximately perform the above operations in tensor version. Refer to Fig. \ref{fig:Algorithm} for easy understanding.

In tensor version, Eq. (\ref{theX}) can be easily performed by the two multiplications between a tensor and a matrix along the $3^{rd}$ mode of the tensor. Specifically, in the TensorFlow\footnote{\url{https://tensorflow.google.cn/}} framework, multiplying $\mathcal{Y}\in\mathbb{R}^{H\times W\times s}$ with matrix $\A\in\mathbb{R}^{s\times S}$ along the channel mode can be easily performed by using the 2D convolution function with a $1\times 1 \times s\times S$ kernel tensor $\mathcal{A}$.
$\hat{\mathcal{Y}}$ and $\B$ can be multiplied similarly. In summary, we can perform the tensor version of (\ref{theX}) by:
\begin{equation}\label{theX2}
  \mathcal{X}^{(k)} = \mathcal{Y}\times_3\A^T+\hat{\mathcal{Y}}^{(k)}\times_3\B^T,
\end{equation}
where $\times_3$ denotes the mode-3 Multiplication for tensor\footnote{For a tensor $\mathcal{U}\in \mathbb{R}^{I\times J \times K}$ with $u_{ijk}$ as its elements, and $\bm{V}\in \mathbb{R}^{K\times L}$ with $v_{kl}$ as its elements, let $\mathcal{W} = \mathcal{U}\times_3\bm{V}$, the elements of $\mathcal{W}$ are $w_{ijl} = \sum_{k=1}^K {u_{ijk}v_{lk}}$. Besides, $\mathcal{W} = \mathcal{U}\times_3\bm{V} \Leftrightarrow \bm{W} = \bm{U}\bm{V}^T$. }.

In Eq. (\ref{theE}), the matrix $\C$ represents the spatial down-sampling operator, which can be
decomposed into 2D convolutions and down-sampling operators \cite{hardie2004map, molina1999bayesian,molina2008variational}.
%For example, in the proposed method we use a 2D channel-wise convolution operator with a $4\times4$ filter and an average pooling operator to reduce image size by 4 times in spatial resolution for a HS image.
%Moreover, for large factor spatial reduction, we use several spatial 4 times spatial reductions and 2 times spatial reductions to approach it.
%For example, in the proposed network, we use two 4 times spatial reductions and a 2 times spatial reductions to approach the 32 times spatial reduction.
Thus, we perform the tensor version of (\ref{theE}) by:
\begin{equation}\label{theE2}
    \mathcal{E}^{(k)} = \mbox{downSample}_{\theta_d^{(k)}}\left(\mathcal{X}^{(k)}\right)-\mathcal{Z},
\end{equation}
where $\mathcal{E}^{(k)}$ is an $h\times w\times S$ tensor, $\mbox{downSample}_{\theta_d^{(k)}}(\cdot)$ is the downsampling network consisting of 2D channel-wise convolutions and average pooling operators, and $\theta_d^{(k)}$ denotes filters involved in the operator at the $k^{th}$ stage of network.

In Eq. (\ref{theG}), the transposed matrix $\C^T$ represents a spatial upsampling operator. This operator can be easily performed by exploiting the 2D transposed convolution \cite{dumoulin2016guide}, which is the transposition of the combination of convolution and downsampling operator. By exploiting the 2D transposed convolution with filter in the same size with the one used in (\ref{theE2}), we can approach (\ref{theG}) in the network by:\vspace{-1mm}
\begin{equation}\label{theG2}
      \mathcal{G}^{(k)} = \eta\cdot\mbox{upSample}_{\theta_u^{(k)}}\left(\mathcal{E}^{(k)}\right)\times_3\B, \vspace{-1mm}
\end{equation}
where $\mathcal{G}^{(k)}\in\mathbb{R}^{H\times W\times S}$, $\mbox{upSample}_{\theta_u^{(k)}}(\cdot)$ is the spacial upsampling network consisting of transposed convolutions and $\theta_u^{(k)}$ denotes the corresponding filters in  the $k^{th}$ stage.

In Eq. (\ref{theY}),  $\mbox{prox}(\cdot)$ is a to-be-decided proximal operator. We adopt the deep residual network (ResNet) \cite{he2016deep} to learn this operator. We then represent (\ref{theY}) in our network as:\vspace{-1mm}
\begin{equation}\label{theY2}
        \hat{\mathcal{Y}}^{(k+1)} = \mbox{proxNet}_{\theta_p^{(k)}}\left(\hat{\mathcal{Y}}^{(k)} - \mathcal{G}^{(k)}\right),\vspace{-1mm}
\end{equation}
where $\mbox{proxNet}_{\theta_p^{(k)}}(\cdot)$ is a ResNet which represents the proximal operator in our algorithm and the parameters involved in the ResNet at the $k^{th}$ stage are denoted by $\theta_p^{(k)}$.

With Eq. (\ref{theX2})-(\ref{theY2}), we can now construct the stages in the proposed network.  Fig. \ref{fig:Network} (b) shows the flowchart of a single stage of the proposed network.

\textbf{Normal stage.} In the first stage, we simply set $\hat{\mathcal{Y}}^{(1)} = \bm{0}$. By exploiting (\ref{theX2})-(\ref{theY2}), we can obtain the first network stage as shown in Fig. \ref{fig:Network} (c). Fig. \ref{fig:Network} (d) shows the $k^{th}$ stage ($1<k<K$) of the network obtained by utilizing (\ref{theX2})-(\ref{theY2}).

\textbf{Final stage.} As shown in Fig. \ref{fig:Network}(e), in the final stage, we can approximately generate the HrHS image by (\ref{theX2}). Note that $\X^{(K)}$ (the unfolding matrix of $\mathcal{X}^{(K)}$) has been intrinsically encoded with low-rank structure. Moreover, according to \textbf{Theorem 1}, there exists an $\R\in \mathbb{R}^{S\times s}$, s.t.,  $\Y = \X^{(K)}\!\R$, which satisfies the observation model (\ref{modelY}).

However,  HrMS images $\mathcal{Y}$ are usually corrupted with slight noise in reality, and there is a little gap between the low rank assumption and the real situation. This implies that $\X^{(K)}$ is not exactly equivalent to the to-be-estimated HrHS image. Therefore, as shown in Fig. \ref{fig:Network} (e),  in the final stage of the network, we add a ResNet on $\mathcal{X}^{(K)}$ to adjust the gap between the to-be-estimated HrHS image and the $\X^{(K)}$: \vspace{-1mm}
\begin{equation}\label{output}
  \hat{\mathcal{X}} = \mbox{resNet}_{\theta_{r}}\left(\mathcal{X}^{(K)}\right).\vspace{-1mm}
\end{equation}

In this way, we design an end-to-end training architecture, dubbed as HSI fusion net. We denote the entire MS/HS fusion net as
$\mathcal{\hat{\mathcal{X}}}= \mbox{MHFnet}\left({\mathcal{Y},\mathcal{Z},\Theta}\right)$,
%\begin{equation}\label{Network}
%  \mathcal{\hat{\mathcal{X}}}= \mbox{MHFnet}\left({\mathcal{Y},\mathcal{Z},\Theta}\right),
%\end{equation}
where $\Theta$ represents all the parameters involved in the network, including $\A$, $\B$,  $\{\theta_d^{(k)},\theta_u^{(k)},\theta_p^{(k)} \}_{k=1}^{K-1}$, $\theta_d^{(K)}$ and $\theta_{r}$. Please refer to supplementary material for more details of the network design.

\subsection{Network training}\vspace{-1mm}
\textbf{Training loss.}
As shown in Fig. \ref{fig:Network} (e), the training loss for each training image is defined as following:
\begin{equation}\label{Loss}
  L = \|\hat{\mathcal{X}}\!-\!\mathcal{X}\|_F^2\!+\!\alpha\!\sum_{k=1}^K\nolimits\| \mathcal{X}^{(k)}\! -\! \mathcal{X} \|_F^2\!+\!\beta\| \mathcal{E}^{(K)} \|_F^2,
\end{equation}
where  $\hat{\mathcal{X}}$ and $\mathcal{X}^{(k)}$ are the final  and per-stage outputs of the proposed network,  $\alpha$ and $\beta$ are two trade-off parameters\footnote{We set $\alpha$ and $\beta$ with small values ($0.1$ and $0.01$, respectively) in all experiments, to make the first term play a dominant role.}.
%Such simple settings make our method perform well consistently throughout all experiments.
The first term is the pixel-wise $L_2$ distance between the output of the proposed network and the ground truth $\mathcal{X}$, which is the main component of our loss function. The second term is the pixel-wise $L_2$ distance between the output $\mathcal{X}^{(k)}$  and the ground truth $\mathcal{X}$ in each stage. This term helps find the correct parameters in each stage, since appropriate $\hat{\mathcal{Y}}^{(k)}$ would lead to $\hat{\mathcal{X}^{(k)}} \approx \mathcal{X} $. The final term is the pixel-wise $L_2$ distance of the residual of observation model (\ref{modelZ}) for the final stage of the network.

  \begin{figure}[t]
  \begin{center}
  \vspace{-3mm}
     \includegraphics[width=1\linewidth]{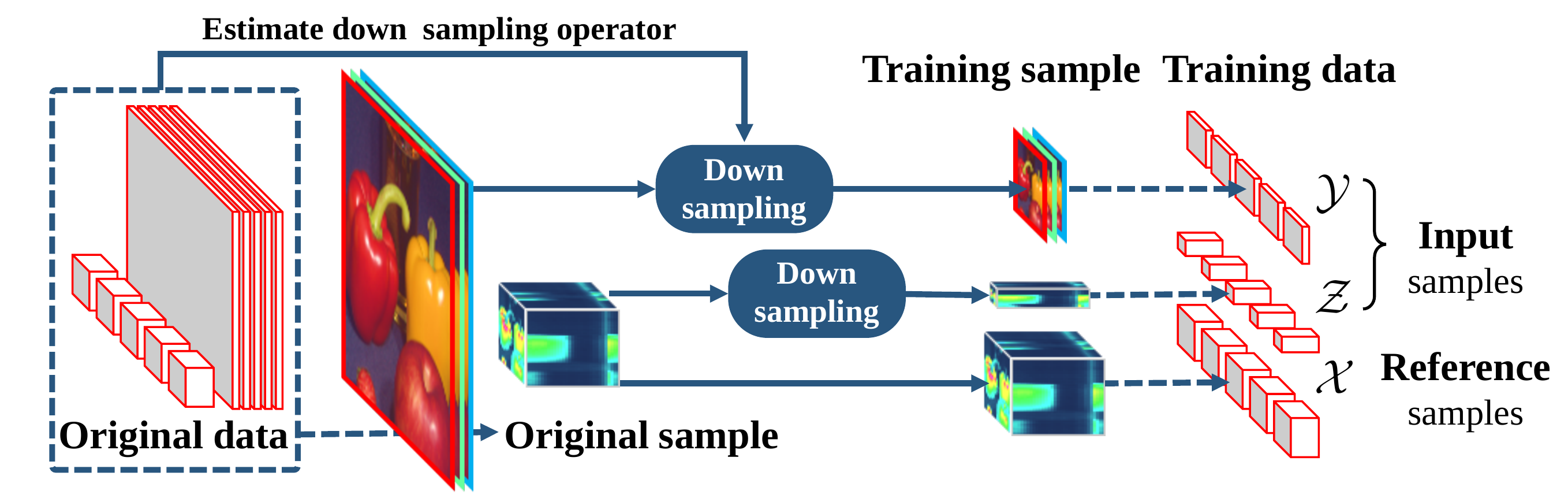}
  \end{center}
  \vspace{-3mm}
     \caption{Illustration of how to create the training data when HrHS images are unavailable.}
  \label{fig:TrainData}
    \vspace{-4mm}
  \end{figure}

\textbf{Training data.} For simulation data and  real data with available ground-truth HrHS images, we can easily use the paired training data $\{ (\mathcal{Y}_n, \mathcal{Z}_n), \mathcal{X}_n\}_{n=1}^N$ to learn the parameters in the proposed MHF-net.
Unfortunately,  for real data, HrHS images $\mathcal{X}_n$s are sometimes unavailable. In this case,  we use the method proposed in \cite{scarpa2018target} to address this problem, where the Wald protocol \cite{zeng2010fusion} is used to create the training data as shown in Fig. \ref{fig:TrainData}.
We downsample both HrMS images and  LrHS images, so that the original LrHS images can be taken as references for the downsampled data. Please refer to supplementary material for more details.
%In order to match the sensor properties, we estimate the spatial  downsampling operator $\C$  with the observed HrMS images and  LrHS images. Due to space limit, please refer to supplementary material for more detail of the  downsampling operator estimation.
%the estimation of the downsampling operator.

\textbf{Implementation details.} %In our method, we easily set the trade-off parameters $\alpha$ and $\beta$ in the loss function as $0.1$ and $0.01$, respectively.
We implement and train our network using TensorFlow
%\footnote{\url{https://tensorflow.google.cn/}}
framework. We use Adam optimizer to train the network for 50000 iterations with a batch size of 10 and a learning rate of 0.0001.  The initializations of the parameters and other implementation details are listed in supplementary materials.
%We initialize the parameter $\A=(\bar{\Y}\bar{\Y})^{-1}\bar{\Y}^T\bar{\X}$, where $\bar{\Y}$ and $\bar{\X}$ are matrices obtained by stacking all the HrMS and HrHS images in the training data along the spatial dimension\footnote{$\A=(\bar{\Y}\bar{\Y})^{-1}\bar{\Y}^T\bar{\X}$ is a solution of $\min_{\A}\| \bar{\Y}\A-\bar{\X}\|_F^2$.}. We initialize the filters in the donwsampling net (\ref{theE2}) and upsampleing net (\ref{theG2}) with $p\times p$ matrices  whose elements are all equal to $\frac{1}{p^2}$, where $p$ is the size of filter. We initialize the other parameters in MHF-net with zero-mean Gaussion distribution with  standard deviation  0.1.

\vspace{-2mm}
\section{Experimental results}\vspace{-1mm}

We first conduct simulated experiments to verify the mechanism of MHF-net quantitatively. Then, experimental results on simulated and real data sets are demonstrated to evaluate the performance of MHF-net.

\textbf{Evaluation measures.} Five quantitative picture quality indices (PQI) are employed for performance evaluation, including peak signal-to-noise ratio
(PSNR), spectral angle mapper (SAM) \cite{yuhas1993determination}, erreur relative globale adimensionnelle
de synth$\grave{e}$se (ERGAS \cite{wald2002data}), structure similarity (SSIM \cite{wang2004image}), feature similarity
(FSIM \cite{zhang2011fsim}).
SAM calculates the average angle between spectrum vectors of the target MSI and the reference one across all spatial positions and ERGAS measures fidelity of the restored image based on the weighted sum of MSE in each band. PSNR, SSIM and FSIM are conventional PQIs. They evaluate the similarity between the target and the reference images based on MSE and structural consistency, perceptual consistency, respectively.
The smaller ERGAS and SAM are, and the larger PSNR, SSIM and FSIM are, the better the fusion result is.

  \begin{figure*}[t]
  \begin{center}
      \vspace{-3mm}
     \includegraphics[width=0.98\linewidth]{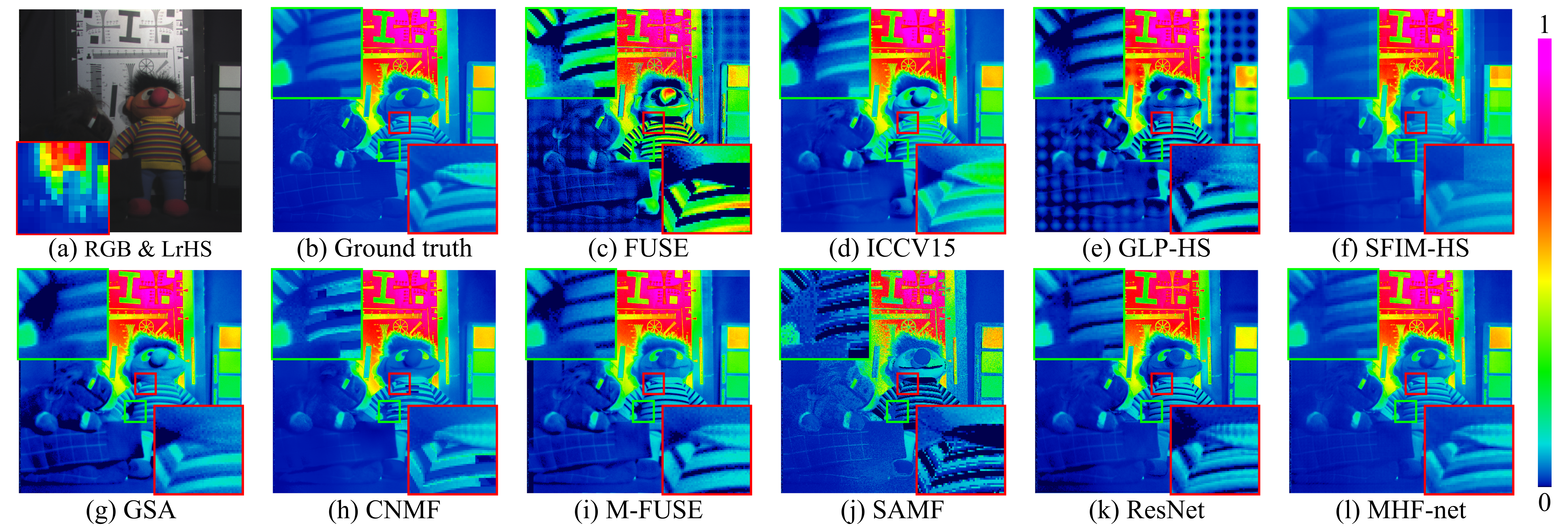}
  \end{center}
  \vspace{-4mm}
     \caption{(a) The simulated RGB (HrMS) and LrHS (left bottom) images of \emph{chart and staffed toy}, where we display the 10th (490nm) band of the HS image. (b) The ground-truth HrHS image. (c)-(l) The results obtained by 10  comparison methods, with two demarcated areas zoomed in 4 times for easy observation.  } %The two demarcated areas are 4 times amplified.
  \label{fig:FIg4}
    \vspace{-4mm}
  \end{figure*}

\subsection{Model verification with CAVE data}\vspace{-1mm}

To verify the efficiency of the proposed MHF-net, we first compare the performance of MHF-net with different settings on the CAVE Multispectral Image Database \cite{yasuma2010generalized}\footnote{\url{http://www.cs.columbia.edu/CAVE/databases/}}.
The database consists of 32 scenes with spatial size of $512\times 512$, including full spectral resolution reflectance data from 400nm to 700nm at 10nm steps (31 bands in total). We generate the HrMS image (RGB image) by integrating all the ground truth HrHS bands with the same simulated spectral response $\R$, and generate the LrHS images via downsampling the ground-truth with a factor of $32$ implemented by averaging over $32\times32$ pixel blocks as \cite{akhtar2014sparse, kawakami2011high}.

To prepare samples for training, we randomly select $20$ HS images from CAVE database and extract $96\times96$ overlapped patches from them as reference HrHS images for training. Then the utilized HrHS, HrMS and LrHS images are of size  ${96\times 96\times 31}$, ${96\times 96\times 3}$  and  ${3\times 3\times 31}$, respectively. The remaining $12$ HS images of the database are used for validation, where the original images are treated as ground truth HrHS images, and the HrMS and LrHS images are generated similarly as the training samples.

We compare the performance of the proposed MHF-net under different stage number $K$. In order to make the competition fair, we adjust the level number $L$ of the ResNet used in $\mbox{proxNet}_{\theta_p^{(k)}}$ for each situation, so that the total level number of the network in each setting is similar to each other.
Moreover, to better verify the efficiency of the proposed network, we implement another network for competition, which only uses the ResNet in (\ref{theY2}) and (\ref{output}) without using other structures in MHF-net. This method is simply denoted as ``ResNet". In this method, we set the input as $[\mathcal{Y},\mathcal{Z}_{up}]$, where $\mathcal{Z}_{up}$ is obtained by  interpolating the LrHS image $\mathcal{Z}$ (using a bicubic filter) to the dimension of $\mathcal{Y}$ as \cite{palsson2017multispectral} did. We set the level number of ResNet to be 30.

\begin{table}
%\vspace{1mm}
\begin{center}
\caption{Average performance of the competing methods over 12 testing samples of CAVE data set with respect to 5 PQIs. }\label{table1}
\vspace{1.5mm}
\footnotesize
\setlength{\tabcolsep}{7.5pt}
\begin{tabular}{cccccc}
  \hline
  \hline
          & \multirow{2}{*}{ResNet}   &\multicolumn{4}{c}{MHF-net with $(K,L)$}  \\
         \cline{3-6}
          &  &    $(4,9)$  &    $(7,5)$   &   $(10,4)$  &    $(13,2)$ \\
    \hline
  PSNR    &     32.25   &    36.15   &    36.61   &    36.85   &   \textbf{  37.23 } \\
   SAM    &    19.093   &    9.206   &    8.636   &    7.587   &  \textbf{  7.298} \\
  ERGA    &    141.28   &    92.94   &    88.56   &    86.53   &   \textbf{  81.87 } \\
  SSIM    &     0.865   &    0.948   &    0.955   &    0.960   &  \textbf{  0.962} \\
  FSIM    &     0.966   &    0.974   &    0.975   &    0.975   &  \textbf{  0.976} \\
  \hline
  \hline
\end{tabular}
\normalsize
\end{center}
\vspace{-8mm}
\end{table}

Table \ref{table1} shows the average results over 12 testing HS images of two DL methods in different settings. We can observe that MHF-net with more stages, even with fewer net levels in total, can significantly lead to better performance. We can also observe that the MHF-net can achieve better results than ResNet (about 5db in PSNR), while the main difference between MHF-net and ResNet is our proposed stage structure in the network. These results show that the proposed stage structure in MHF-net, which introduces interpretability specifically to the problem, can indeed help enhance the performance of MS/HS fusion.

\subsection{Experiments with simulated data}\vspace{-1mm}
We then evaluate MHF-net on simulated data in comparison with state-of-art methods.

\textbf{Comparison methods.} The comparison methods include: FUSE \cite{wei2015fast}\footnote{\scriptsize{\url{http://wei.perso.enseeiht.fr/publications.html}}}, ICCV15 \cite{lanaras2015hyperspectral}\footnote{\scriptsize{\url{https://github.com/lanha/SupResPALM}}}, GLP-HS \cite{selva2015hyper}\footnote{\scriptsize{\url{http://openremotesensing.net/knowledgebase/hyperspectral-and-multispectral-data-fusion/}}\label{fn:repeat}}, SFIM-HS \cite{liu2000smoothing}\footref{fn:repeat}, GSA \cite{aiazzi2007improving}\footref{fn:repeat}, CNMF \cite{yokoya2011coupled}\footnote{\scriptsize{\url{http://naotoyokoya.com/Download.html}}}, M-FUSE \cite{wei2016blind}\footnote{\scriptsize{\url{https://github.com/qw245/BlindFuse}}} and SASFM \cite{huang2014spatial}\footnote{We write the code by ourselves.}, representing the state-of-the-art traditional methods. We also compare the proposed MHF-net with the implemented ResNet method.

  \begin{figure*}[t]
    \vspace{-3mm}
  \begin{center}
     \includegraphics[width=0.95\linewidth]{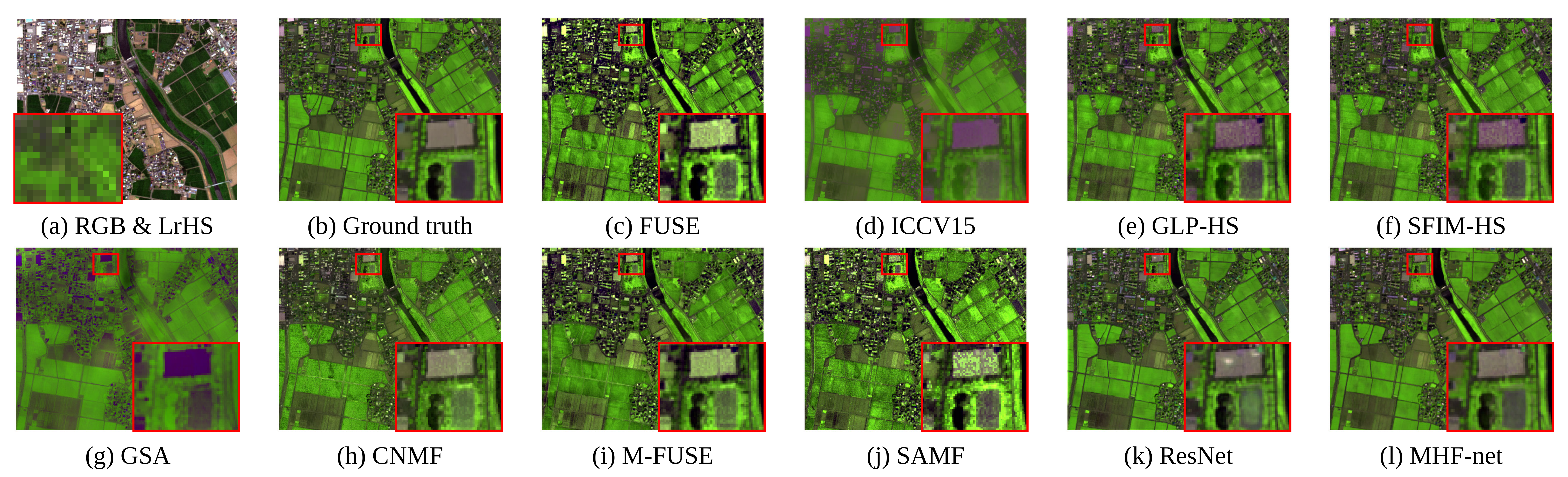}
  \end{center}
   \vspace{-4mm}
     \caption{(a) The simulated RGB (HrMS) and LrHS (left bottom)  images of a test sample in Chikusei data set. We show the composite image of the HS image with bands 70-100-36  as R-G-B. (b) The ground-truth HrHS image. (c)-(l) The results obtained by 10  comparison methods, with a demarcated area zoomed in 4 times for easy observation. }
  \label{fig:FIg5}
    \vspace{-2mm}
  \end{figure*}

  \begin{figure*}[t]
  \begin{center}
     \includegraphics[width=0.94\linewidth]{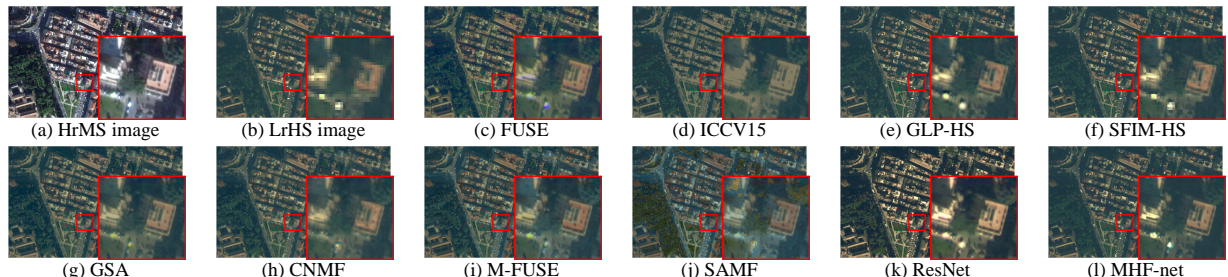}
  \end{center}
  \vspace{-4mm}
     \caption{(a) and (b) are  the HrMS (RGB) and  LrHS images of the left bottom area of \emph{Roman Colosseum} acquired by World View-2 (WV-2). We show the composite image of the HS image with bands 5-3-2  as R-G-B. (c)-(l) The results obtained by 10 comparison methods, with a demarcated area zoomed in 5 times for easy observation. }%, with a demarcated area zoomed in 5 times for easy observation
  \label{fig:FIg6}
    \vspace{-4mm}
  \end{figure*}

\begin{table}
%\vspace{1mm}
\begin{center}
\caption{Average performance  of the competing methods over 12 testing images of CAVE date set with respect to 5 PQIs. }\label{table2}
\vspace{1.5mm}
\footnotesize
\setlength{\tabcolsep}{7.5pt}
\begin{tabular}{cccccc}
  \hline
  \hline
&   PSNR &   SAM &  ERGAS &   SSIM &   FSIM \\
    \hline
  FUSE    &     30.95   &    13.07   &   188.72   &    0.842   &     0.933 \\
ICCV15    &     32.94   &    10.18   &   131.94   &    0.919   &     0.961 \\
GLP-HS    &     33.07   &    11.58   &   126.04   &    0.891   &     0.942 \\
SFIM-HS   &     31.86   &     7.63   &   147.41   &    0.914   &     0.932 \\
   GSA    &     33.78   &    11.56   &   122.50   &    0.884   &     0.959 \\
  CNMF    &     33.59   &     8.22   &   122.12   &    0.929   &     0.964 \\
M-FUSE    &     32.11   &     8.82   &   151.97   &    0.914   &     0.947 \\
 SASFM    &     26.59   &    11.25   &   362.70   &    0.799   &     0.916 \\
ResNet    &     32.25   &    16.14   &   141.28   &    0.865   &     0.966 \\
MHF-net    &   \textbf{  37.23}   &  \textbf{   7.30}   &   \textbf{ 81.87}   &   \textbf{ 0.962}   &    \textbf{ 0.976} \\
  \hline
  \hline
\end{tabular}
\normalsize
\end{center}
\vspace{-9mm}
\end{table}

\textbf{Performance comparison with CAVE data.} With the same experiment setting as previous section, we compare the performance of all competing methods on the 12 testing HS images ($K=13$ and $L=2$ in MHF-net). Table \ref{table2} lists the average performance over all testing images of all comparison methods. From the table, it is seen that the proposed MHF-net method can significantly outperform other competing methods with respect to all evaluation measures. Fig. \ref{fig:FIg4}  shows the $10$-th band (490nm) of the HS image \emph{chart and staffed toy} obtained by the completing methods. It is easy to observe that the proposed method performs better than other competing ones, in the better recovery of both finer-grained textures and coarser-grained structures.
More results are depicted in the supplementary material.

\textbf{Performance comparison with Chikusei data.} The Chikusei data set \cite{yokoya2017hyperspectral}\footnote{\url{http://naotoyokoya.com/Download.html}} is an airborne HS image taken over Chikusei, Ibaraki, Japan, on 29 July 2014. The data set is of size $2517\times 2335\times128 $ with the spectral range from 0.36 to 1.018. We view the original data as the HrHS image and  simulate the HrMS (RGB image) and LrMS (with a factor of 32) image in the similar way as the previous section.

We select a $500\times 2210$-pixel-size image from the top area of the original data for training, and extract $96\times96$ overlapped patches from the training data as reference HrHS images for training.
The input HrHS, HrMS and LrHS samples are of sizes  ${96\times 96\times 128}$, ${96\times 96\times 3}$  and  ${3\times 3\times 128}$, respectively.
Besides, from remaining part of the original image, we extract 16 non-overlap $448\times544\times128$ images as testing data. More details about the experimental setting are introduced in supplementary material.

Table \ref{table3} shows the average performance over 16 testing images of all competing methods. It is easy to observe that the proposed method significantly outperforms other methods with respect to all evaluation measures. Fig. \ref{fig:FIg5} shows the composite images of a test sample obtained by the competing methods, with bands 70-100-36  as R-G-B. It is seen that the composite image obtained by MHF-net is closest to the ground-truth, while the results of other methods usually contain obvious incorrect structure or spectral distortion. More results are listed in supplementary material.

\begin{table}
%\vspace{1mm}
\begin{center}
\caption{Average performance of the competing methods over 16 testing samples of Chikusei data set  with respect to 5 PQIs. }\label{table3}
\vspace{1.5mm}
\footnotesize
\setlength{\tabcolsep}{7.5pt}
\begin{tabular}{cccccc}
  \hline
  \hline
&   PSNR &   SAM &  ERGAS &   SSIM &   FSIM \\
    \hline
  FUSE     &     26.59   &     7.92   &   272.43   &    0.718   &     0.860 \\
ICCV15     &     27.77   &     3.98   &   178.14   &    0.779   &     0.870 \\
GLP-HS     &     28.85   &     4.17   &   163.60   &    0.796   &     0.903 \\
SFIM-HS    &     28.50   &     4.22   &   167.85   &    0.793   &     0.900 \\
   GSA     &     27.08   &     5.39   &   238.63   &    0.673   &     0.835 \\
  CNMF     &     28.78   &     3.84   &   173.41   &    0.780   &     0.898 \\
 M-FUSE    &     24.85   &     6.62   &   282.02   &    0.642   &     0.849 \\
 SASFM     &     24.93   &     7.95   &   369.35   &    0.636   &     0.845 \\
ResNet     &     29.35   &     3.69   &   144.12   &    0.866   &     0.930 \\
MHF-net    &   \textbf{  32.26}   &  \textbf{   3.02}   &   \textbf{109.55}   &   \textbf{ 0.890}   &    \textbf{ 0.946} \\

  \hline
  \hline
\end{tabular}
\normalsize
\end{center}
\vspace{-9mm}
\end{table}

\subsection{Experiments with real data}\vspace{-1mm}
%completing methods: GSA\cite{aiazzi2007improving}, CNMF \cite{yokoya2011coupled}
In this section, sample images of \emph{Roman Colosseum} acquired by World View-2 (WV-2) are used in our experiments\footnote{\url{https://www.harrisgeospatial.com/DataImagery/SatelliteImagery/HighResolution/WorldView-2.aspx}}. This data set contains an HrMS image (RGB image) of size $1676\times 2632 \times 3$ and an LrHS image of size  $419\times 658 \times 8$, while the HrHS image is not available. We select the top half part of the HrMS ($836\times 2632 \times 3$) and LrHS ($209\times 658 \times 8$) image to train the MHF-net, and exploit the remaining parts of the data set as testing data. We first extract the training data into $144\times144\times3$ overlapped HrMS patches and $36\times36\times3$ overlapped LrHS patches and then generate the training samples by the method as shown in Fig. \ref{fig:TrainData}. The input HrHS, HrMS and LrHS samples are of size  ${36\times 36\times 8}$, ${36\times 36\times 3}$  and  ${9\times 9\times 8}$, respectively.

Fig. \ref{fig:FIg5} shows a portion of the fusion result of the testing data (left bottom area of the original image). Visual inspection evidently shows that the proposed method gives the better visual effect.
By comparing with the results of ResNet, we can find that  the results of both methods are clear, but the color and brightness of result of the proposed method are much closer to the LrHS image.

%\vspace{-1mm}
\section{Conclusion}%\vspace{-1mm}
In this paper, we have provided a new MS/HS fusion network.
The network takes the advantage of deep learning that all parameters can be learned from the training data with fewer prior pre-assumptions on data, and furthermore takes into account the generation mechanism underlying the MS/HS fusion data. This is achieved by constructing a new MS/HS fusion model based on the observation models, and unfolding the algorithm into an optimization-inspired deep network.
The network is thus specifically interpretable to the task, and can help discover the spatial and spectral response operators in a purely end-to-end manner.
Experiments implemented on simulated and real MS/HS fusion cases have substantiated the superiority of the proposed MHF-net over the state-of-the-art methods.

%-------------------------------------------------------------------------

%{\small
%\bibliographystyle{ieee}
%\bibliography{egbib}
%}

\end{document}